%% file: main.tex
\documentclass[11pt,letterpaper]{article}
\usepackage{emnlp2016}

\usepackage{times}
\usepackage{latexsym}

\emnlpfinalcopy

\def\emnlppaperid{1002}

\usepackage{color,soul} 

\usepackage{graphicx}

\usepackage{multirow}
\usepackage{booktabs}

\usepackage{amsmath}
\usepackage{amsfonts}

\usepackage{hyperref}

\usepackage{pgfplots}
\usetikzlibrary{calc}
\usetikzlibrary{decorations.pathmorphing}
\usetikzlibrary{shapes,arrows,chains}
\tikzset{>=latex}

\usepackage{txfonts}

\usepackage[normalem]{ulem}
\DeclareMathOperator*{\argmax}{arg\,max}

\newcommand{\dtw}{\textrm{DTW}}
\newcommand{\dba}{\textrm{DBA}}
\newcommand{\fastalign}{\texttt{fast\char`_align}}

\let\citet\newcite
\let\citep\cite

\makeatletter
\def\citewith[#1]{\def\citename##1{#1##1, }\@internalcite}
\makeatother

\title{An Unsupervised Probability Model for Speech-to-Translation Alignment
of Low-Resource Languages}

\author{Antonios Anastasopoulos\\
    University of Notre Dame\\
    $\mathtt{aanastas@nd.edu}$\\
    \And
    David Chiang\\
    University of Notre Dame \\
    $\mathtt{dchiang@nd.edu}$\\
    \And
    Long Duong\\
    University of Melbourne \\
    $\mathtt{lduong@student.unimelb.edu.au}$
}

\date{}

\begin{document}

\maketitle

\begin{abstract}
For many low-resource languages, spoken language resources are more likely to be annotated with translations than with transcriptions. Translated speech data is potentially valuable for documenting endangered languages or for training speech translation systems. A first step towards making use of such data would be to automatically align spoken words with their translations.
We present a model that combines Dyer et al.'s reparameterization of IBM Model 2 (\fastalign) and $k$-means clustering using Dynamic Time Warping as a distance metric. The two components are trained jointly using expectation-maximization. In an extremely low-resource scenario, our model performs significantly better than both a neural model and a strong baseline.
\end{abstract}

\section{Introduction}

For many low-resource languages, speech data is easier to obtain than textual data. And because speech transcription is a costly and slow process, speech is more likely to be annotated with translations than with transcriptions. This translated speech is a potentially valuable source of information -- for example, for documenting endangered languages or for training speech translation systems. 

In language documentation, data is usable only if it is interpretable.
To make a collection of speech data usable for future studies of the language, something resembling interlinear glossed text (transcription, morphological analysis, word glosses, free translation) would be needed at minimum. New technologies are being developed to facilitate collection of translations \cite{bird+al:2014}, and there already exist recent examples of parallel speech collection efforts focused on endangered languages \cite{blachon2016parallel,adda2016breaking}. As for the other annotation layers, one might hope that a first pass could be done automatically. A first step towards this goal would be to automatically align spoken words with their translations, capturing information similar to that captured by word glosses. 

In machine translation, statistical models have traditionally required alignments between the source and target languages as the first step of training. Therefore, producing alignments between speech and text would be a natural first step towards MT systems operating directly on speech.

We present a model that, in order to learn such alignments, adapts and combines two components: Dyer et al.'s reparameterization of IBM Model 2 \cite{dyer2013simple}, more commonly known as \verb|fast_align|, and $k$-means clustering using Dynamic Time Warping \cite{berndt1994using} as a distance metric.
The two components are trained jointly using expectation-maximization. 

We experiment on two language pairs. One is Spanish-English, using the CALLHOME and Fisher corpora. The other is Griko-Italian; Griko is an endangered language for which we created (and make freely available)\footnote{\url{https://www3.nd.edu/~aanastas/griko/griko-data.tar.gz}} gold-standard translations and word alignments \cite{grikodatabase}. In all cases, our model outperforms both a naive but strong baseline and a neural model \cite{long-EtAl:2016:NAACL-HLT}.

\section{Background}

In this section, we briefly describe the existing models that the two components of our model are based on. In the next section, we will describe how we adapt and combine them to the present task.

\subsection{IBM Model 2 and \fastalign}

The IBM translation models \cite{brown1993mathematics} aim to model the distribution $p(\mathbf{e} \mid \mathbf{f})$ for an English sentence \mbox{$\mathbf{e} = e_1\cdots e_l$}, given a French sentence $\mathbf{f} = f_1\cdots e_m$. They all introduce a hidden variable $\mathbf{a} = a_1 \cdots a_l$ that gives the position of the French word to which each English word is aligned. 

The general form of IBM Models 1, 2 and \fastalign{} is
\begin{align*}
p(\mathbf{e},\mathbf{a} \mid \mathbf{f}) = p(l) \prod_{i=1}^{l}t(e_i \mid f_{a_i}) \, \delta(a_i\mid i,l,m)
\end{align*}
where $t(e \mid f)$ is the probability of translating French word $f$ to English word $e$, and $\delta(a_i=j \mid i, l, m)$ is the probability of aligning the $i$-th English word with the $j$-th French word.

In Model 1, $\delta$ is uniform; in Model 2, it is a categorical distribution. \citet{dyer2013simple} propose a reparameterization of Model 2, known as \fastalign:
\begin{align*}
    h(i,j,l,m) &= - \left\vert \frac{i}{l} - \frac{j}{m} \right\vert \\
        \delta(a_i \mid i,l,m) &= \begin{cases}
             p_0 & a_i=0  \\
             (1-p_0) \frac{\exp \lambda h(i,a_i,l,m)}{Z_{\lambda}(i,l,m)} & a_i > 0 
\end{cases}
\end{align*}
where the null alignment probability $p_0$ and precision $\lambda \geq 0$ are hyperparameters optimized by grid search.
As $\lambda \rightarrow 0$, the distribution gets closer to the distribution of IBM Model 1, and as $\lambda$ gets larger, the model prefers monotone word alignments more strongly.

\subsection{DTW and DBA}

Dynamic Time Warping (DTW) \cite{berndt1994using} is a dynamic programming method for measuring distance between two temporal sequences of variable length, as well as computing an alignment based on this distance. 
Given two sequences $\phi,\phi'$ of length $m$ and $m'$ respectively, \dtw{} constructs an $m\times m'$ matrix $w$.
The warping path can be found by evaluating the following recurrence:
\[
w_{i,j} = d(\phi_i,\phi'_j) + \min\lbrace w_{i-1,j}, w_{i-1,j-1}, w_{i,j-1} \rbrace
\]
where $d$ is a distance metric.
In this paper, we normalize the cost of the warping path:
\[\dtw{(\phi,\phi')}=\frac{w_{m,m'}}{m+m'}\] 
which lies between zero and one.

\dtw{} Barycenter Averaging (DBA) \cite{petitjean2011global} is an iterative approximate method that attempts to find a centroid of a set of sequences, minimizing the sum of squared \dtw{} distances.

In the original definition, given a set of sequences, DBA chooses one sequence randomly to be a ``skeleton.'' Then, at each iteration, \dba{} computes the \dtw{} between the skeleton and every sequence in the set, aligning each of the skeleton's points with points in all the sequences. The skeleton is then refined using the found alignments, by updating each frame in the skeleton to the mean of all the frames  aligned to it. In our implementation, in order to avoid picking a skeleton that is too short or too long, we randomly choose one of the sequences with median length.

\section{Model}
\label{sec:Model}

We use a generative model from a source-language speech segment consisting of feature frames $\boldsymbol{\phi} = \phi_1 \cdots \phi_m$ to a target-language segment consisting of words $\mathbf{e} = e_1 \ldots e_l$.
We chose to model $p(\mathbf{e} \mid \boldsymbol{\phi})$ rather than $p(\boldsymbol{\phi} \mid \mathbf{e})$ 
because it makes it easier to incorporate \dtw. The other direction is also possible, and we plan to explore it in future work.

In addition to the target-language sentence $\mathbf{e}$, our model hypothesizes a sequence $\textbf{f} = f_1\cdots f_l$ of source-language clusters (intuitively, source-language words), and spans $(a_i,b_i)$ of the source signal that each target word $e_i$ is aligned to. Thus, the clusters $\mathbf{f} = f_1 \cdots f_l$ and the spans $\mathbf{a} = a_1, \ldots, a_l$ and $\mathbf{b} = b_1, \ldots, b_l$ are the hidden variables of the model:
\[
p(\mathbf{e} \mid \boldsymbol{\phi}) = \sum_{\mathbf{a}, \mathbf{b}, \mathbf{f}} p(\mathbf{e}, \mathbf{a}, \mathbf{b}, \mathbf{f} \mid \boldsymbol{\phi}).
\]

The model generates $\mathbf{e}, \mathbf{a}, \mathbf{b}$, and $\mathbf{f}$ from $\boldsymbol{\phi}$ as follows.
\begin{enumerate}
\item Choose $l$, the number of target words, with uniform probability. (Technically, this assumes a maximum target sentence length, which we can just set to be very high.)
\item For each target word position $i=1, \ldots, l$:
\begin{enumerate}
\item Choose a cluster $f_i$.
\item Choose a span of source frames $(a_i, b_i)$ for $e_i$ to be aligned to. \label{step2b}
\item Generate a target word $e_i$ from $f_i$.
\end{enumerate}
\end{enumerate}
Accordingly, we decompose $p(\mathbf{e}, \mathbf{a}, \mathbf{b}, \mathbf{f} \mid \boldsymbol{\phi})$ into several submodels:
\begin{align*}
p(\mathbf{e}, \mathbf{a}, \mathbf{b}, \mathbf{f} \mid \boldsymbol{\phi}) = p(l) \prod_{i=1}^l \,
 &u(f_i) \times {} \\
 & s(a_i, b_i \mid f_i, \boldsymbol{\phi}) \times {}\\
&\delta(a_i,b_i \mid  i,l,|\boldsymbol{\phi}|) \times {}\\
 &t(e_i \mid f_i).
\end{align*}
Note that submodels $\delta$ and $s$ both generate spans (corresponding to step \ref{step2b}), making the model deficient. We could make the model sum to one by replacing $u(f_i) s(a_i, b_i \mid f_i, \boldsymbol{\phi})$ with $s(f_i \mid a_i, b_i, \boldsymbol{\phi})$, and this was in fact our original idea, but the model as defined above works much better, as discussed in Section \ref{sec:prop-vs-def}. We describe both $\delta$ and $s$ in detail below.

\paragraph{Clustering model} The probability over clusters, $u(f)$, is just a categorical distribution.
The submodel $s$ assumes that, for each cluster $f$, there is a ``prototype'' signal $\boldsymbol{\phi}^f$ \citewith[cf.~]{ristad1998learning}. Technically, the $\boldsymbol{\phi}^f$ are parameters of the model, and will be recomputed during the M step. 
Then we can define:
\[
s(a, b \mid f, \boldsymbol{\phi}) = \frac{\exp(-\dtw (\boldsymbol{\phi}^f, \phi_a \cdots \phi_b)^2)}{\sum_{a,b=1}^m\exp(-\dtw (\boldsymbol{\phi}^f, \phi_a \cdots \phi_b)^2)}
\]
where $\dtw$ is the distance between the prototype and the segment computed using Dynamic Time Warping. Thus $s$ assigns highest probability to spans of $\boldsymbol{\phi}$ that are most similar to the prototype $\boldsymbol{\phi}^f$.

\begin{figure}
\begin{center}
\includegraphics[scale=0.27]{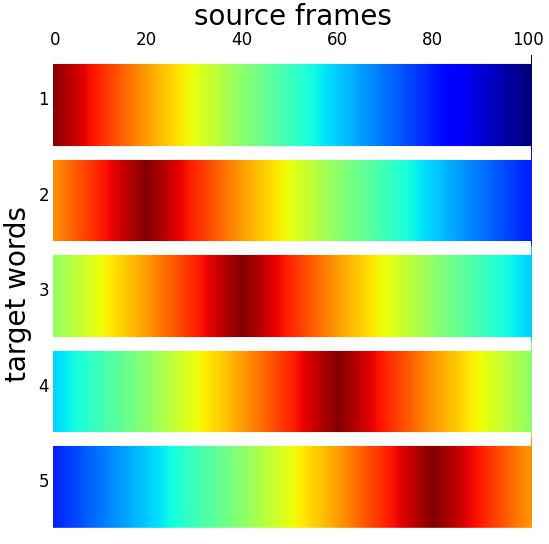}
\includegraphics[scale=0.27]{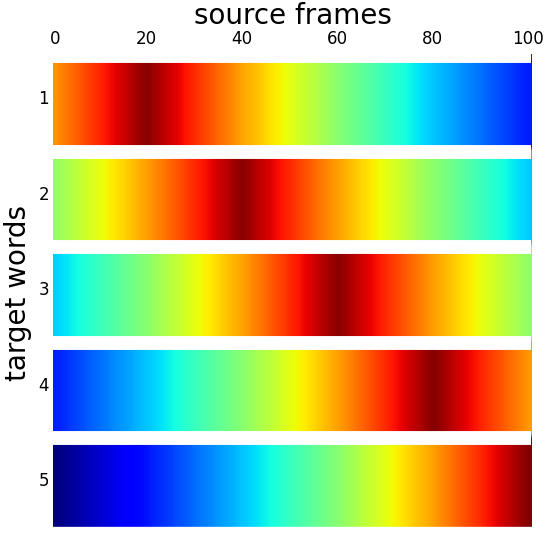}
\end{center}
\caption{Sample distributions for the alignment variables~$a$ and~$b$ for $m=100$, $l=5$, $p_0=0$, $\lambda=0.5$, and $\mu=20$.}
\label{fig:dists}
\end{figure}

\paragraph{Distortion model}
The submodel $\delta$ controls the reordering of the 
target words relative to the 
source frames.
It is an adaptation of \fastalign{} to our setting, where there is not a single source
word position $a_i$, but a span $(a_i, b_i)$. We want the model to prefer the middle of the word to be close to the diagonal, so we need the variable $a$ to be somewhat to the left and $b$ to be somewhat to the right. 
Therefore, we introduce an additional hyperparameter $\mu$ which is intuitively the number of frames in a word.
Then we define
\begin{align*}
    h_a(i,j,l,m,\mu) &= - \left\vert \frac{i}{l} - \frac{j}{m-\mu} \right\vert \\
    h_b(i,j,l,m,\mu) &= 
    - \left\vert \frac{i}{l} - \frac{j-\mu}{m-\mu} \right\vert \\
    \delta_a(a_i \mid i,l,m) &= \begin{cases} p_0 & a_i=0  \\
             (1-p_0) \frac{\exp \lambda h_a(i,a_i,l,m)}{Z_{\lambda}(i,l,m)} & a_i > 0 
        \end{cases} \\
    \delta_b(b_i \mid i,l,m) &= \begin{cases} p_0 & b_i=0  \\
             (1-p_0) \frac{\exp \lambda h_b(i,b_i,l,m)}{Z_{\lambda}(i,l,m)} & b_i > 0 
        \end{cases} \\
        \delta(a_i,b_i \mid i, l, m) &= \delta_a(a_i \mid i, l, m) \, \delta_b(b_i \mid i, l, m)
\end{align*}
where the $Z_\lambda(i,l,m)$ are set so that all distributions sum to one.
Figure~\ref{fig:dists} shows an example visualisation of the 
the resulting distributions for the two variables of our model.
    
We set $\mu$ differently for each word. For each $i$, we set $\mu_i$ to be proportional to the number of \emph{characters} in $e_i$, such that $\sum_i \mu_i = m$.

\paragraph{Translation model} The translation model $t(e\mid f)$ is just a categorical distribution, in principle allowing a many-to-many relation between source clusters and target words. To speed up training (with nearly no change in accuracy, in our experiments), we restrict this relation so that there are $k$ source clusters for each target word, and a source cluster uniquely determines its target word. Thus, $t(e\mid f)$ is fixed to either zero or one, and does not need to be reestimated. In our experiments, we set $k=2$, allowing each target word to have up to two source-language translations/pronunciations. (If a source word has more than one target translation, they are treated as distinct clusters with distinct prototypes.)

\section{Training}

We use the hard (Viterbi) version of the Expectation-Maximization (EM) algorithm to estimate the parameters of our model, because calculating expected counts in full EM would be prohibitively expensive, requiring  summations over all possible alignments.

Recall that the hidden variables of the model are the alignments ($a_i, b_i$) and the source words ($f_i$). The parameters are the translation probabilities $t(e_i \mid f)$ and the prototypes ($\boldsymbol\phi^f$). 
The (hard) E step uses the current model and prototypes to find, for each target word, the best source segment to align it to and the best source word.
The M step reestimates the probabilities \mbox{$t(e \mid f)$} and the prototypes $\boldsymbol\phi^f$.
We describe each of these steps in more detail below.

\paragraph{Initialization}

Initialization is especially important since we are using hard EM. 

To initialize the parameters, we initialize the hidden variables and then perform an M step.
We associate each target word type $e$ with $k=2$ source clusters, and for each occurrence of $e$, we randomly assign it one of the $k$ source clusters.

The alignment variables $a_i,b_i$ are initialized to
\[
a_i,b_i = \argmax_{a,b} \delta(a,b \mid i, l, m).
\] 

\paragraph{M step}
The M step reestimates the probabilities $t(e \mid f)$ using relative-frequency estimation. 

The prototypes $\boldsymbol\phi^f$ are more complicated. Theoretically, the M step should recompute each $\boldsymbol\phi^f$ so as to maximize that part of the log-likelihood that depends on $\boldsymbol\phi^f$:
\begin{align*}
L_{\boldsymbol\phi^f} &= \sum_{\boldsymbol\phi} \sum_{i\mid f_i = f} \log s(a_i, b_i \mid f, \boldsymbol\phi) \\
 &= \sum_{\boldsymbol\phi} \sum_{i\mid f_i = f} \log \frac{\exp (-\dtw(\phi^f, \phi_{a_i} \cdots \phi_{b_i})^2)}{Z(f,\boldsymbol\phi)}\\
 &= \sum_{\boldsymbol\phi} \sum_{i\mid f_i = f}
-\dtw(\phi^f, \phi_{a_i} \cdots \phi_{b_i})^2 - \log Z(f,\boldsymbol\phi)
\end{align*}
where the summation over $\boldsymbol\phi$ is over all source signals in the training data.
This is a hard problem, but note that the first term is just the sum-of-squares of the \dtw{} distance between $\boldsymbol\phi^f$ and all source segments that are classified as $f$. This is what \dba{} is supposed to approximately minimize, so we simply set $\boldsymbol\phi^f$ using \dba{}, ignoring the denominator.

\paragraph{E step}
The (hard) E step uses the current model and prototypes to find, for each target word, the best source segment to align it to and the best source cluster.

In order to reduce the search space for $\mathbf{a}$ and $\mathbf{b}$, we use the unsupervised phonetic boundary detection method of \citet{khanagha:hal-01059348}. This method operates directly on the speech signal and provides us with candidate phone boundaries, on which we restrict the possible values for $\mathbf{a}$ and $\mathbf{b}$, creating a list of candidate utterance spans.

Furthermore, we use a simple silence detection method. We pass the envelope of the signal through a low-pass filter, and then mark as ``silence'' time spans of 50ms or longer in which the magnitude is below a threshold of 5\% relative to the maximum of the whole signal. This method is able to detect about 80\% of the total pauses, with a 90\% precision in a 50ms window around the correct silence boundary. We can then remove from the candidate list the utterance spans that include silence, on the assumption that a word should not include silences. Finally, in case one of the span's boundaries happens to be within a silence span, we also move it so as to not include the silence. 

\paragraph{Hyperparameter tuning}

The hyperparameters $p_0$, $\lambda$, and $\mu$ are not learned. We simply set $p_0$ to zero (disallowing unaligned target words) and set $\mu$ as described above.

For $\lambda$ 
we perform a grid search over candidate values to maximize the alignment F-score on the development set. We obtain the best scores with~$\lambda=0.5$.

\section{Related Work}

A first step towards modelling parallel speech can be performed by modelling phone-to-word alignment, instead of directly working on continuous speech. For example, \citet{stahlberg2012word} extend IBM Model 3 to align phones to words in order to build cross-lingual pronunciation lexicons. Pialign \cite{neubig2012substring} aligns characters and can be applied equally well to phones. \citet{long-EtAl:2016:NAACL-HLT} use an extension of the neural attentional model of \citet{DBLP:journals/corr/BahdanauCB14} for aligning phones to words and speech to words; we discuss this model below in Section~\ref{sec:baselines}.

There exist several supervised approaches that attempt to integrate speech recognition and machine translation. However, they rely heavily on the abundance of training data, pronunciation lexicons, or language models, and therefore cannot be applied in a low- or zero-resource setting.

A task somewhat similar to ours, which operates at a monolingual level, is the task of zero-resource spoken term discovery, which aims to discover repeated words or phrases in continuous speech. Various approaches \cite{ten2007computational,park2008unsupervised,muscariello2009audio,zhang2010towards,jansen2010towards} have been tried, in order to spot keywords, using segmental DTW to identify repeated trajectories in the speech signal. 

\citet{kamperunsupervised} try to discover word segmentation and a pronunciation lexicon in a zero-resource setting, combining \dtw{} with acoustic embeddings; their methods operate in a very low-vocabulary setting. \citet{Bansal:Thesis:2015} attempts to build a speech translation system in a  low-resource setting, by using as source input the simulated output of an unsupervised term discovery system.

\section{Experiments}

We evaluate our method on two language pairs, Spanish-English and Griko-Italian, against two baseline methods, a naive baseline, and the model of \citet{long-EtAl:2016:NAACL-HLT}.

\subsection{Data}

For each language pair, we require a sentence-aligned parallel corpus of source-language speech and target-language text. A subset of these sentences should be annotated with span-to-word alignments for use as a gold standard.

\subsubsection{Spanish-English}
\label{sec:spa-en-exp}

For Spanish-English, we use the Spanish CALLHOME corpus (LDC96S35) and the Fisher corpus (LDC2010T04), which consist of telephone conversations between Spanish native speakers based in the US and their relatives abroad, together with English translations produced by~\citet{post2013improved}. Spanish is obviously not a low-resource language, but we pretend that it is low-resource by not making use of any Spanish ASR or resources like transcribed speech or pronunciation lexicons.

Since there do not exist gold standard alignments between the Spanish speech and English words, we use the ``silver'' standard alignments produced by \citet{long-EtAl:2016:NAACL-HLT} for the CALLHOME corpus, and followed the same procedure for the Fisher corpus as well. In order to obtain them, they first used a forced aligner
to align the speech to its transcription, and GIZA++ with the \texttt{gdfa} symmetrization heuristic
to align the Spanish transcription to the English translation. They then combined the two alignments to produce ``silver'' standard alignments between the Spanish speech and the English words.

The CALLHOME dataset consists of~17532 Spanish utterances, based on the dialogue turns.
We first use a sample of~2000 sentences, out of which we use~200 as a development set and the rest as a test set. We also run our experiments on the whole dataset, selecting~500 utterances for a development set, using the rest as a test set. 
The Fisher dataset consists of~143355 Spanish utterances. We use~1000 of them as a development set and the rest as a test set.

\subsubsection{Griko-Italian}

We also run our model on a corpus that consists of about~20 minutes of speech in Griko, an endangered minority dialect of Greek spoken in south Italy, along with text translations into Italian~\cite{grikodatabase}.\footnote{\url{http://griko.project.uoi.gr/}} The corpus consists of~330 mostly prompted utterances by nine native speakers.
Although the corpus is very small, 
we use it to showcase the effectiveness of our method in a hard setting with extremely low resources.

All utterances were manually annotated and transcribed by a trained linguist and bilingual speaker of both languages, who produced the Griko transcriptions and Italian glosses. We created full translations into Italian and manually aligned the translations with the Griko transcriptions. We then combined the two alignments (speech-to-transcription and transcription-to-translation) to produce speech-to-translation alignments. Therefore, our comparison is done against an accurate ``gold'' standard alignment. 
We split the data into a development set of just~30 instances, and a test set of the remaining~300 instances. 

\subsubsection{Preprocessing}

In both data settings, we treat the speech data as a sequence of~39-dimensional Perceptual Linear Prediction (PLP) vectors encoding the power spectrum of the speech signal~\cite{hermansky1990perceptual}, computed at $10$ms intervals. We also normalize the features at the utterance level, shifting and scaling them to have zero mean and unit variance.

\subsection{Baselines}
\label{sec:baselines}

Our naive baseline assumes that there is no reordering between the source and target language, and aligns each target word $e_i$ to a source span whose length in frames is proportional to the length of $e_i$ in characters. This actually performs very well on language pairs that show minimal or no reordering, and language pairs that have shared or related vocabularies.

The other baseline that we compare against is the neural network attentional model of \citet{long-EtAl:2016:NAACL-HLT}, which extends the attentional model of \citet{DBLP:journals/corr/BahdanauCB14} to be used for aligning and translating speech, and, along with several modifications, achieve good results on the phone-to-word alignment task, and almost match the baseline performance on the speech-to-word alignment task.

\section{Results}

To evaluate an automatic alignment between the speech and its translation against the gold/silver standard alignment, we compute alignment precision, recall, and F-score as usual, but on links between source-language frames and target-language words.

\begin{table}
\input{tables/results-table2.tex}
\end{table}

\subsection{Overview}

Table~\ref{tab:results} shows the precision, recall, and balanced F-score of the three models on the Spanish-English CALLHOME corpus (both the 2000-sentence subset and the full set), the Spanish-English Fisher corpus, and the Griko-Italian corpus. 

In all cases, our model outperforms both the naive baseline and the neural attentional model.
Our model, when compared to the baselines, improves greatly on precision, while slightly underperforming the naive baseline on recall. In certain applications, higher precision may be desirable: for example, in language documentation, it's probably better to err on the side of precision; in phrase-based translation, higher-precision alignments lead to more extracted phrases.

The rest of the section provides a further analysis of the results, focusing on the extremely low-resource Griko-Italian dataset.

\subsection{Speaker robustness}
\begin{figure*}[t]
\begin{center}
\resizebox{6.5in}{!}{
\input{examples/speaker-example-2.tex}
}
\caption{Alignments produced for the Italian sentence \texttt{devo comprare il pane ogni giorno} as uttered by three different Griko speakers.}
\label{fig:speaker}
\end{center}
\end{figure*}
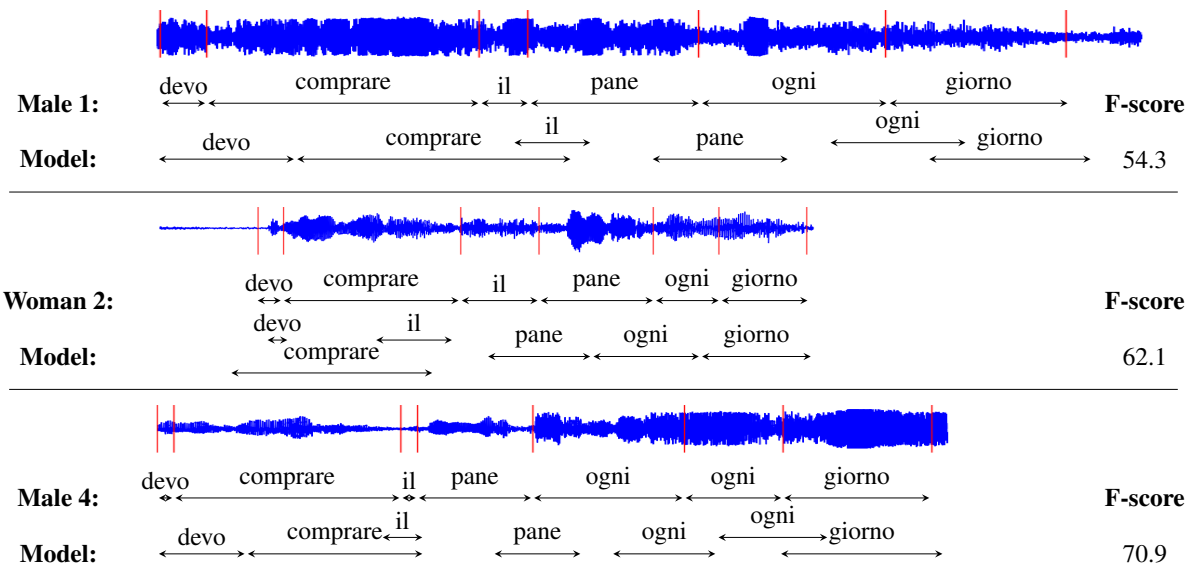

\begin{table}
\input{tables/speaker-table.tex}
\end{table}

Figure~\ref{fig:speaker} shows the alignments produced by our model for three utterances of the same sentence from the Griko-Italian dataset by three different speakers. Our model's performance is roughly consistent across these utterances. In general, the model does not seem significantly affected by speaker-specific variations, as shown in Table~\ref{tab:speaker}. 

We do find, however, that the performance on male speakers is slightly higher compared to the female speakers. 
This might be because the female speakers' utterances are, on average, longer by about~2 words than the ones uttered by males.

\subsection{Word level analysis}

\begin{figure}[ht]
\begin{center}
\resizebox{3in}{!}{
\input{tikzpic/word-level-hist-10.tex}
}
\caption{There is a positive correlation between average word-level F-score and average word utterance length (in frames).}
\label{fig:word-level-f1}
\end{center}
\end{figure}
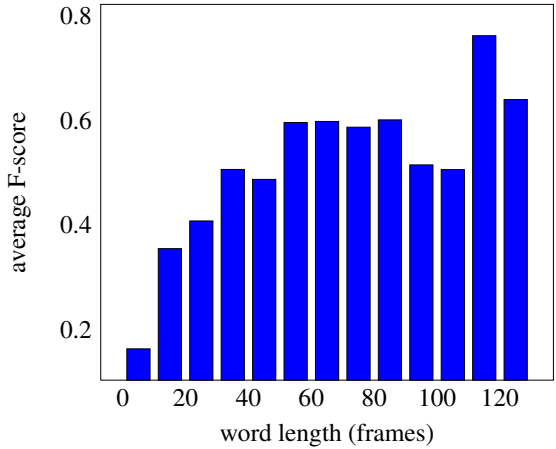

We also compute F-scores for each Italian word type. As shown in Figure~\ref{fig:word-level-f1}, the longer the word's utterance, the easier it is for our model to correctly align it. Longer utterances seem to carry enough information for our \dtw-based metric to function properly.
On the other hand, shorter utterances are harder to align. The vast majority of Griko utterances that have less than~20 frames and are less accurately aligned correspond to monosyllabic determiners (\texttt{o}, \texttt{i},\texttt{a}, \texttt{to}, \texttt{ta}) or conjunctions and prepositions (\texttt{ka}, \texttt{ce}, \texttt{en}, \texttt{na}, \texttt{an}). For such short utterances, there could be several parts of the signal that possibly match the prototype, leading the clustering component to prefer to align to wrong spans.

Furthermore, we note that rare word types tend to be correctly aligned. The average F-score for hapax legomena (on the Italian side) is~$63.2$, with~$53\%$ of them being aligned with an F-score higher than~$70.0$.

\subsection{Comparison with proper model}
\label{sec:prop-vs-def}

\begin{figure*}
\begin{minipage}{.48\textwidth}
\resizebox{\textwidth}{!}{
\input{examples/example2.tex}
}
\caption{The deficient model performs very well, whereas the proper and the attentional model prefer extreme alignment spans. For example, the proper model's alignment for the words  \texttt{dovevo} and \texttt{pane} are much too short.}
\label{fig:Griko_alignment1}
\end{minipage}
\hfill
\begin{minipage}{.48\textwidth}
\begin{flushright}  
\resizebox{\textwidth}{!}{
\input{examples/example1.tex}
}
\caption{One of the rare examples where the proper model performs better than the deficient one. The hapax legomena \texttt{Valeria} and \texttt{giornali} are not properly handled by the attentional model.}
\label{fig:Griko_alignment2}
\end{flushright}
\end{minipage}
\end{figure*}

As mentioned in Section~\ref{sec:Model}, our model is deficient, but it performs much better than the model that sums to one (henceforth, the ``proper'' model): In the Spanish-English dataset~(2000 sentences sample) the proper model yields an F-score of~32.1, performing worse than the naive baseline; in the Griko-Italian dataset, it achieves an F-score of~44.3, which is better than the baselines, but still worse than our model.

In order to further examine why this happens, we performed three EM iterations on the Griko-Italian dataset with our model (in our experience, three iterations are usually enough for convergence)%
, and then computed one more E step with both our model and the proper model, so as to ensure that the two models would align the dataset using the exact same prototypes and that their outputs will be comparable.

In this case, the proper model achieved an overall F-score of~44.0, whereas our model achieved an F-score of~53.6. 
Figures~\ref{fig:Griko_alignment1} and~\ref{fig:Griko_alignment2} show the resulting alignments for two sentences. In both of these examples, it is clear that the proper model prefers extreme spans: the selected spans are either much too short or (less frequently) much too long. This is further verified by examining the statistics of the alignments: the average span selected by the proper model has a length of about~$30\pm39$ frames whereas the average span of the alignments produced by our deficient model is~$37\pm24$ frames. This means that the alignments of the deficient model are much closer to the gold ones, whose average span is~$42\pm26$ frames.

We think that this is analogous to the ``garbage collection'' problem in word alignment. In the IBM word alignment models, if a source word $f$ occurs in only one sentence, then EM can align many target words to $f$ and learn a very peaked distribution $t(e \mid f)$. This can happen in our model and the proper model as well, of course, since IBM Model~2 is embedded in them. But in the proper model, something similar can also happen with $s(f \mid a, b)$: EM can make the span $(a, b)$ large or small, and evidently making the span small allows it to learn a very peaked distribution $s(f \mid a, b)$. By contrast, our model has $s(a, b \mid f)$,  which seems less susceptible to this kind of effect.

\section{Conclusion}

Alignment of speech to text translations is a relatively new task, one with particular relevance for low-resource or endangered languages. The model we propose here, which combines \fastalign{} and $k$-means clustering using \dtw{} and \dba{}, outperforms both a very strong naive baseline and a neural attentional model, on three tasks of various sizes. 

The language pairs that we work on do not have very much word reordering, and more divergent language pairs should prove more challenging. In that case, the naive baseline should be much less competitive. Similarly, the \fastalign-based distortion model may become less appopriate; we could turn the hyperparameter $\lambda$ down, bringing it closer to IBM Model 1, but we plan to try incorporating IBM Model 3 or the HMM alignment model \cite{vogel1996hmm} instead.

Finally, we will investigate downstream applications of our alignment methods, in the areas of both language documentation and speech translation.

\bibliography{References}
\bibliographystyle{emnlp2016}

\end{document}

%% file: tables/results-table2.tex
\begin{center}
\begin{tabular}{{@{}lll|lll@{}}}
\toprule
& & method & precision & recall & F-score \\ 
\midrule
\multirow{6}*{\rotatebox{90}{\parbox{3cm}{\centering CALLHOME spa-eng}}} &
\multirow{3}*{\rotatebox{0}{\parbox{.6cm}{\centering 2k sents}}} & ours & \textbf{38.8} & 38.9 & \textbf{38.8} \\
& &naive & 31.9 & \textbf{40.8} & 35.8\\
& &neural & 23.8 & 29.8 & 26.4 \\
\cmidrule{2-6}
& \multirow{3}{*}{\rotatebox{0}{\parbox{.6cm}{\centering 17k sents}}}  & ours & \textbf{38.4} & 38.8 & \textbf{38.6} \\
& &naive  & 31.8 & \textbf{40.7} & 35.7\\
& &neural & 26.1 & 32.9 & 29.1 \\

\midrule
\multirow{3}{*}{\rotatebox{90}{\parbox{1.5cm}{\centering Fisher spa-eng}}} &
\multirow{3}{*}{\rotatebox{0}{\parbox{.6cm}{\centering 143k sents}}} & ours & \textbf{33.3} & 28.7 & \textbf{30.8} \\
& &naive  & 24.0 & \textbf{33.2} & 27.8\\
& & & & &\\

\midrule 
\multirow{3}{*}{\rotatebox[origin=c]{90}{gri-ita}}& \multirow{3}{*}{\rotatebox{0}{\parbox{.6cm}{\centering 300 sents}}} &
ours & \textbf{56.6} & 51.2 & \textbf{53.8} \\
& &naive & 42.2 & \textbf{52.2} & 46.7 \\
& &neural & 24.6 & 30.0 & 27.0 \\
\bottomrule
\end{tabular}
\end{center}
\caption{Our model achieves higher precision and F-score than both the naive baseline and the neural model on all datasets.}
\label{tab:results}

%% file: examples/speaker-example-2.tex
\begin{tikzpicture}[scale=0.5]

\pgfmathsetmacro{\SPA}{13.7}
\pgfmathsetmacro{\MA}{12}
\pgfmathsetmacro{\SPB}{7.7}
\pgfmathsetmacro{\MB}{6}
\pgfmathsetmacro{\SPC}{1.7}
\pgfmathsetmacro{\MC}{0}

\node[inner sep=0pt] at (15.8,\SPA+2){\includegraphics[width=16.4cm,height=1cm]{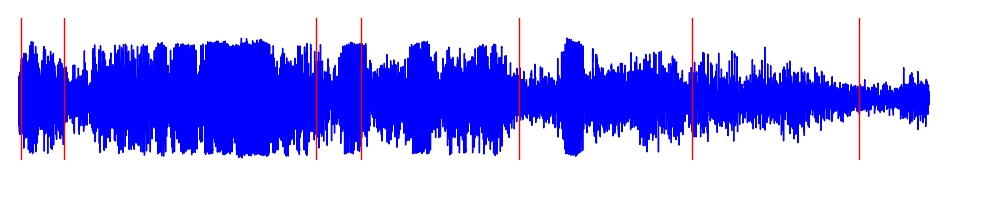}};
\node[align=left] at (-3.0,\SPA) (0) {\textbf{Male 1:}};
\node[inner sep=0pt] at (0.1,\SPA) (G1) {};
\node[inner sep=0pt] at (1.5,\SPA) (G2) {};
\node[inner sep=0pt] at (9.8,\SPA) (G3) {};
\node[inner sep=0pt] at (11.3,\SPA) (G4) {};
\node[inner sep=0pt] at (16.5,\SPA) (G5) {};
\node[inner sep=0pt] at (22.2,\SPA) (G6) {};
\node[inner sep=0pt] at (27.7,\SPA) (G7) {};
\node[inner sep=0pt] at (30.0,\SPA) (sc) {\textbf{F-score}};

\draw[<->, >=stealth] (G1) -- node [midway,above,align=center] {devo} (G2);
\draw[<->, >=stealth] (G2) -- node [midway,above,align=center] {comprare} (G3);
\draw[<->, >=stealth] (G3) -- node [midway,above,align=center] {il} (G4);
\draw[<->, >=stealth] (G4) -- node [midway,above,align=center] {pane} (G5);
\draw[<->, >=stealth] (G5) -- node [midway,above,align=center] {ogni} (G6);
\draw[<->, >=stealth] (G6) -- node [midway,above,align=center] {giorno} (G7);

\node[align=left] at (-3.0,\MA) (0) {\textbf{Model:}};
\node[inner sep=0pt] at (0.0,\MA) (a1) {};
\node[inner sep=0pt] at (4.2,\MA) (a2) {};
\node[inner sep=0pt] at (4.2,\MA) (b1) {};
\node[inner sep=0pt] at (12.6,\MA) (b2) {};
\node[inner sep=0pt] at (10.8,\MA+.5) (c1) {};
\node[inner sep=0pt] at (13.2,\MA+.5) (c2) {};
\node[inner sep=0pt] at (15.0,\MA) (d1) {};
\node[inner sep=0pt] at (19.2,\MA) (d2) {};
\node[inner sep=0pt] at (20.4,\MA+.5) (e1) {};
\node[inner sep=0pt] at (24.6,\MA+.5) (e2) {};
\node[inner sep=0pt] at (23.4,\MA) (f1) {};
\node[inner sep=0pt] at (28.4,\MA) (f2) {};
\node[inner sep=0pt] at (30.0,\MA) (sc1) {54.3};

\draw[<->, >=stealth] (a1) -- node [midway,above,align=center] {devo} (a2);
\draw[<->, >=stealth] (b1) -- node [midway,above,align=center] {comprare} (b2);
\draw[<->, >=stealth] (c1) -- node [midway,above,align=center] {il} (c2);
\draw[<->, >=stealth] (d1) -- node [midway,above,align=center] {pane} (d2);
\draw[<->, >=stealth] (e1) -- node [midway,above,align=center] {ogni} (e2);
\draw[<->, >=stealth] (f1) -- node [midway,above,align=center] {giorno} (f2);

\draw (-4.5,\MA-1) -- (31,\MA-1);

\node[inner sep=0pt] at (10.2,\SPB+2){\includegraphics[width=10.5cm,height=1cm]{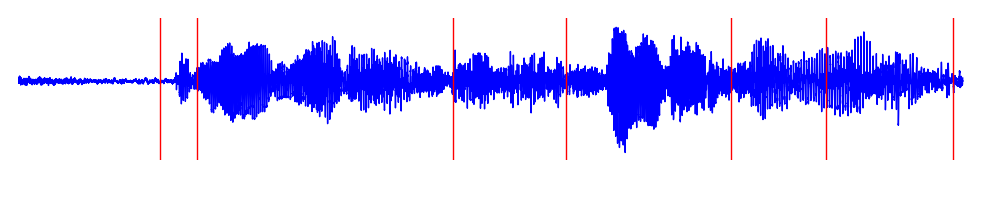}};

\node[align=left] at (-3.0,\SPB) (0) {\textbf{Woman 2:}};
\node[inner sep=0pt] at (3.0,\SPB) (G12) {};
\node[inner sep=0pt] at (3.8,\SPB) (G22) {};
\node[inner sep=0pt] at (9.2,\SPB) (G32) {};
\node[inner sep=0pt] at (11.6,\SPB) (G42) {};
\node[inner sep=0pt] at (15.1,\SPB) (G52) {};
\node[inner sep=0pt] at (17.1,\SPB) (G62) {};
\node[inner sep=0pt] at (19.8,\SPB) (G72) {};
\node[inner sep=0pt] at (30.0,\SPB) (sc22) {\textbf{F-score}};

\draw[<->, >=stealth] (G12) -- node [midway,above,align=center] {devo} (G22);
\draw[<->, >=stealth] (G22) -- node [midway,above,align=center] {comprare} (G32);
\draw[<->, >=stealth] (G32) -- node [midway,above,align=center] {il} (G42);
\draw[<->, >=stealth] (G42) -- node [midway,above,align=center] {pane} (G52);
\draw[<->, >=stealth] (G52) -- node [midway,above,align=center] {ogni} (G62);
\draw[<->, >=stealth] (G62) -- node [midway,above,align=center] {giorno} (G72);

\node[align=left] at (-3.0,\MB) (0) {\textbf{Model:}};
\node[inner sep=0pt] at (3.3,\MB+.5) (a12) {};
\node[inner sep=0pt] at (4.0,\MB+.5) (a22) {};
\node[inner sep=0pt] at (2.2,\MB-.5) (b12) {};
\node[inner sep=0pt] at (8.4,\MB-.5) (b22) {};
\node[inner sep=0pt] at (6.6,\MB+.5) (c12) {};
\node[inner sep=0pt] at (9.0,\MB+.5) (c22) {};
\node[inner sep=0pt] at (10.0,\MB) (d12) {};
\node[inner sep=0pt] at (13.2,\MB) (d22) {};
\node[inner sep=0pt] at (13.2,\MB) (e12) {};
\node[inner sep=0pt] at (16.5,\MB) (e22) {};
\node[inner sep=0pt] at (16.5,\MB) (f12) {};
\node[inner sep=0pt] at (19.9,\MB) (f22) {};
\node[inner sep=0pt] at (30.0,\MB) (sc12) {62.1};

\draw[<->, >=stealth] (a12) -- node [midway,above,align=center] {devo} (a22);
\draw[<->, >=stealth] (b12) -- node [midway,above,align=center] {comprare} (b22);
\draw[<->, >=stealth] (c12) -- node [midway,above,align=center] {il} (c22);
\draw[<->, >=stealth] (d12) -- node [midway,above,align=center] {pane} (d22);
\draw[<->, >=stealth] (e12) -- node [midway,above,align=center] {ogni} (e22);
\draw[<->, >=stealth] (f12) -- node [midway,above,align=center] {giorno} (f22);

\draw (-4.5,\MB-1) -- (31,\MB-1);

\node[inner sep=0pt] at (13.6,\SPC+2){\includegraphics[width=14.1cm,height=1cm]{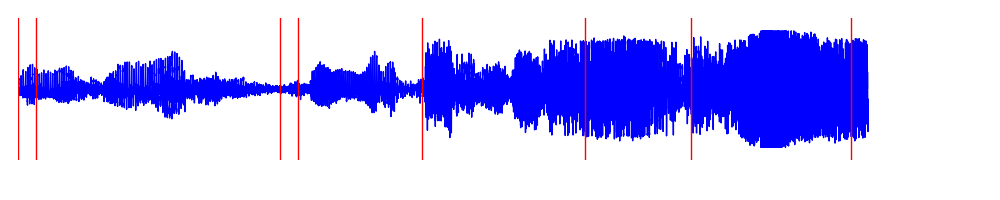}};

\node[align=left] at (-3.0,\SPC) (0) {\textbf{Male 4:}};
\node[inner sep=0pt] at (0.0,\SPC) (G13) {};
\node[inner sep=0pt] at (0.5,\SPC) (G23) {};
\node[inner sep=0pt] at (7.4,\SPC) (G33) {};
\node[inner sep=0pt] at (7.9,\SPC) (G43) {};
\node[inner sep=0pt] at (11.4,\SPC) (G53) {};
\node[inner sep=0pt] at (16.0,\SPC) (G63) {};
\node[inner sep=0pt] at (19.0,\SPC) (G73) {};
\node[inner sep=0pt] at (23.5,\SPC) (G83) {};
\node[inner sep=0pt] at (30.0,\SPC) (sc3) {\textbf{F-score}};

\draw[<->, >=stealth] (G13) -- node [midway,above,align=center] {devo} (G23);
\draw[<->, >=stealth] (G23) -- node [midway,above,align=center] {comprare} (G33);
\draw[<->, >=stealth] (G33) -- node [midway,above,align=center] {il} (G43);
\draw[<->, >=stealth] (G43) -- node [midway,above,align=center] {pane} (G53);
\draw[<->, >=stealth] (G53) -- node [midway,above,align=center] {ogni} (G63);
\draw[<->, >=stealth] (G63) -- node [midway,above,align=center] {ogni} (G73);
\draw[<->, >=stealth] (G73) -- node [midway,above,align=center] {giorno} (G83);

\node[align=left] at (-3.0,\MC) (0) {\textbf{Model:}};
\node[inner sep=0pt] at (0.0,\MC) (a13) {};
\node[inner sep=0pt] at (2.7,\MC) (a23) {};
\node[inner sep=0pt] at (2.7,\MC) (b13) {};
\node[inner sep=0pt] at (8.1,\MC) (b23) {};
\node[inner sep=0pt] at (6.8,\MC+.5) (c13) {};
\node[inner sep=0pt] at (8.1,\MC+.5) (c23) {};
\node[inner sep=0pt] at (10.2,\MC) (d13) {};
\node[inner sep=0pt] at (12.9,\MC) (d23) {};
\node[inner sep=0pt] at (13.8,\MC) (e13) {};
\node[inner sep=0pt] at (17.0,\MC) (e23) {};
\node[inner sep=0pt] at (17.0,\MC+0.5) (f13) {};
\node[inner sep=0pt] at (20.4,\MC+0.5) (f23) {};
\node[inner sep=0pt] at (18.9,\MC) (g13) {};
\node[inner sep=0pt] at (23.9,\MC) (g23) {};
\node[inner sep=0pt] at (30.0,\MC) (sc13) {70.9};

\draw[<->, >=stealth] (a13) -- node [midway,above,align=center] {devo} (a23);
\draw[<->, >=stealth] (b13) -- node [midway,above,align=center] {comprare} (b23);
\draw[<->, >=stealth] (c13) -- node [midway,above,align=center] {il} (c23);
\draw[<->, >=stealth] (d13) -- node [midway,above,align=center] {pane} (d23);
\draw[<->, >=stealth] (e13) -- node [midway,above,align=center] {ogni} (e23);
\draw[<->, >=stealth] (f13) -- node [midway,above,align=center] {ogni} (f23);
\draw[<->, >=stealth] (g13) -- node [midway,above,align=center] {giorno} (g23);

\end{tikzpicture}

%% file: tables/speaker-table.tex
\begin{center}
\begin{tabular}{{@{}l|cc|l@{}}}
\toprule
speaker & utt & len & F-score \\ 
\midrule
female 1 & 55 & 9.0 & 49.4 \\
female 2 & 61 & 8.1 & 55.0 \\
female 3 & 41 & 9.6 & 51.0 \\ 
female 4 & 23 & 7.3 & 54.4 \\
female 5 & 21 & 6.1 & 56.6 \\
\midrule
male 1 & 35 & 5.9 & 59.5 \\
male 2 & 32 & 6.0 & 61.9 \\
male 3 & 34 & 6.7 & 60.2 \\
male 4 & 23 & 6.4 & 64.0 \\
\bottomrule
\end{tabular}
\end{center}
\caption{Model performance (F-score) is generally consistent across speakers. The second column (utt) shows the number of utterances per speaker; the third (len), their average length in words.}
\label{tab:speaker}

%% file: tikzpic/word-level-hist-10.tex
\begin{tikzpicture}
\begin{axis}[
    ytick style={draw=none},
    xtick style={draw=none},
    xlabel = {word length (frames)},
    ylabel = {average F-score}
    ]

    \addplot[ybar,fill=blue] coordinates {
( 5 , 0.163869767683 )
( 15 , 0.355047685314 )
( 25 , 0.407849126836 )
( 35 , 0.50611021423 )
( 45 , 0.487129368078 )
( 55 , 0.595593201029 )
( 65 , 0.5976251304 )
( 75 , 0.586695669005 )
( 85 , 0.600648807056 )
( 95 , 0.514764549208 )
( 105 , 0.505777275568 )
( 115 , 0.761238732581 )
( 125 , 0.639356005735 )
    };
\end{axis}
\end{tikzpicture}

%% file: examples/example2.tex
\begin{tikzpicture}[scale=0.5]
\node[align=left] at (-10,2.5) (gr0) {\textbf{Griko:}};
\node[inner sep=0pt] at (-8.6,2) (gr1) {};
\node[inner sep=0pt] at (-4.3,2) (gr2) {};
\node[inner sep=0pt] at (-3.53,2) (gr3) {};
\node[inner sep=0pt] at (2.19,2) (gr4) {};
\node[inner sep=0pt] at (3.05,2) (gr5) {};
\node[inner sep=0pt] at (6.5,2) (gr6) {};
\draw[<->, >=stealth] (gr1) -- node [midway,above,align=center] {\`{i}cha} (gr2);
\draw[<->, >=stealth] (gr2) -- node [midway,above,align=center] {na} (gr3);
\draw[<->, >=stealth] (gr3) -- node [midway,above,align=center] {afor\`{a}so} (gr4);
\draw[<->, >=stealth] (gr4) -- node [midway,above,align=center] {to} (gr5);
\draw[<->, >=stealth] (gr5) -- node [midway,above,align=center] {tsom\`{i}} (gr6);

\node[inner sep=0pt] at (0,0){\includegraphics[width=9cm]{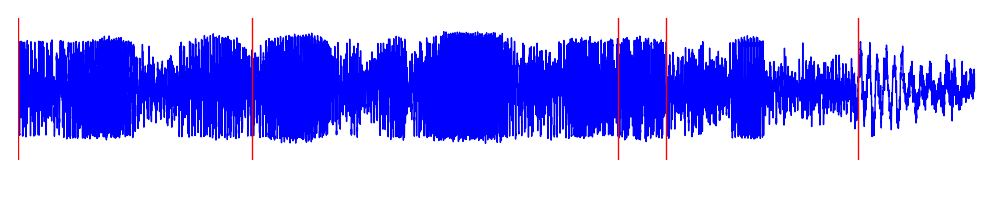}};
\node[align=left] at (-10,-1.5) (0) {\textbf{Gold:}};
\node[inner sep=0pt] at (-8.6,-2) (A) {};
\node[inner sep=0pt] at (-4.3,-2) (B) {};
\node[inner sep=0pt] at (2.19,-2) (C) {};
\node[inner sep=0pt] at (3.05,-2) (D) {};
\node[inner sep=0pt] at (6.5,-2) (E) {};
\node[inner sep=0pt] at (9,-2) (sc) {\textbf{F-score}};
\draw[<->, >=stealth] (A) -- node [midway,above,align=center] {dovevo} (B);
\draw[<->, >=stealth] (B) -- node [midway,above,align=center] {comprare} (C);
\draw[<->, >=stealth] (C) -- node [midway,above,align=center] {il} (D);
\draw[<->, >=stealth] (D) -- node [midway,above,align=center] {pane} (E);
\node[align=left] at (-10,-3.5) (0w) {\textbf{Ours:}};
\node[inner sep=0pt] at (-8.6,-3.5) (A1) {};
\node[inner sep=0pt] at (-3.44,-3.5) (A2) {};
\node[inner sep=0pt] at (-5.15,-4) (B1) {};
\node[inner sep=0pt] at (1.24,-4) (B2) {};
\node[inner sep=0pt] at (1.24,-3.5) (C1) {};
\node[inner sep=0pt] at (2.96,-3.5) (C2) {};
\node[inner sep=0pt] at (4.68,-3.5) (D1) {};
\node[inner sep=0pt] at (7.1,-3.5) (D2) {};
\node[inner sep=0pt] at (9,-3.5) (sc1) {82.3};
\draw[<->, >=stealth] (A1) -- node [midway,above,align=center] {dovevo} (A2);
\draw[<->, >=stealth] (B1) -- node [midway,above,align=center] {comprare} (B2);
\draw[<->, >=stealth] (C1) -- node [midway,above,align=center] {il} (C2);
\draw[<->, >=stealth] (D1) -- node [midway,above,align=center] {pane} (D2);
\node[align=left] at (-10,-5.5) (0w) {\textbf{Proper:}};
\node[inner sep=0pt] at (-4.56,-5.5) (F1) {};
\node[inner sep=0pt] at (-4.2,-5.5) (F2) {};
\node[inner sep=0pt] at (-6.3,-6) (G1) {};
\node[inner sep=0pt] at (2.96,-6) (G2) {};
\node[inner sep=0pt] at (2.39,-5.5) (H1) {};
\node[inner sep=0pt] at (2.96,-5.5) (H2) {};
\node[inner sep=0pt] at (5.92,-5.5) (I1) {};
\node[inner sep=0pt] at (6.5,-5.5) (I2) {};
\node[inner sep=0pt] at (9,-5.5) (sc2) {61.7};

\draw[<->, >=stealth] (F1) -- node [midway,above,align=center] {dovevo} (F2);
\draw[<->, >=stealth] (G1) -- node [midway,above,align=center] {comprare} (G2);
\draw[<->, >=stealth] (H1) -- node [midway,above,align=center] {il} (H2);
\draw[<->, >=stealth] (I1) -- node [midway,above,align=center] {pane} (I2);
\node at (-10,-7.5) (0w) {\textbf{Attention:}};
\node[inner sep=0pt] at (-8.6,-8) (Q1) {};
\node[inner sep=0pt] at (-1.72,-8) (Q2) {};
\node[inner sep=0pt] at (-1.72,-7.5) (W1) {};
\node[inner sep=0pt] at (-0.95,-7.5) (W2) {};
\node[inner sep=0pt] at (-0.95,-8) (E1) {};
\node[inner sep=0pt] at (-0.19,-8) (E2) {};
\node[inner sep=0pt] at (-0.19,-7.5) (R1) {};
\node[inner sep=0pt] at (8.2,-7.5) (R2) {};
\node[inner sep=0pt] at (9,-7.5) (sc3) {38.3};

\draw[<->, >=stealth] (Q1) -- node [midway,above,align=center] {dovevo} (Q2);
\draw[<->, >=stealth] (W1) -- node [midway,above,align=center] {comprare} (W2);
\draw[<->, >=stealth] (E1) -- node [midway,above,align=center] {il} (E2);
\draw[<->, >=stealth] (R1) -- node [midway,above,align=center] {pane} (R2);


\end{tikzpicture}

%% file: examples/example1.tex
\begin{tikzpicture}[scale=0.5]
\node[inner sep=0pt] at (0,0){\includegraphics[width=9cm]{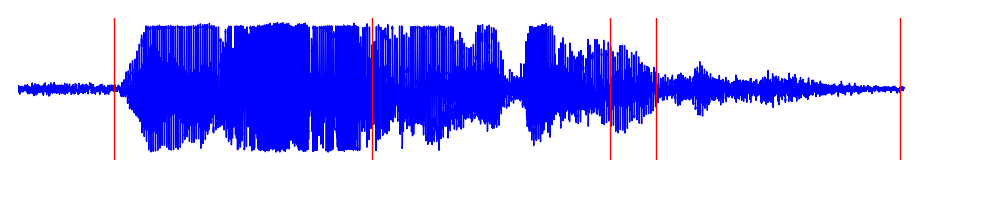}};
\node[align=left] at (-10,2.5) (gr0) {\textbf{Griko:}};
\node[inner sep=0pt] at (-6.88,2) (gr1) {};
\node[inner sep=0pt] at (-6.13,2) (gr2) {};
\node[inner sep=0pt] at (-2.25,2) (gr3) {};
\node[inner sep=0pt] at (1.99,2) (gr4) {};
\node[inner sep=0pt] at (2.82,2) (gr5) {};
\node[inner sep=0pt] at (7.2,2) (gr6) {};
\draw[<->, >=stealth] (gr1) -- node [midway,above,align=center] {\`{e}} (gr2);
\draw[<->, >=stealth] (gr2) -- node [midway,above,align=center] {Val\`{e}ria} (gr3);
\draw[<->, >=stealth] (gr3) -- node [midway,above,align=center] {melet\`{a}} (gr4);
\draw[<->, >=stealth] (gr4) -- node [midway,above,align=center] {\`{o}} (gr5);
\draw[<->, >=stealth] (gr5) -- node [midway,above,align=center] {giorn\`{a}li} (gr6);

\node[align=left] at (-10,-1.5) (0) {\textbf{Gold:}};
\node[inner sep=0pt] at (-6.88,-2) (A) {};
\node[inner sep=0pt] at (-2.25,-2) (B) {};
\node[inner sep=0pt] at (1.99,-2) (C) {};
\node[inner sep=0pt] at (2.82,-2) (D) {};
\node[inner sep=0pt] at (7.2,-2) (E) {};
\node[inner sep=0pt] at (9,-2) (sc) {\textbf{F-score}};
\draw[<->, >=stealth] (A) -- node [midway,above,align=center] {Valeria} (B);
\draw[<->, >=stealth] (B) -- node [midway,above,align=center] {legge} (C);
\draw[<->, >=stealth] (C) -- node [midway,above,align=center] {il} (D);
\draw[<->, >=stealth] (D) -- node [midway,above,align=center] {giornale} (E);
\node[align=left] at (-10,-3.5) (0w) {\textbf{Ours:}};
\node[inner sep=0pt] at (-6.95,-3.5) (A1) {};
\node[inner sep=0pt] at (-3.59,-3.5) (A2) {};
\node[inner sep=0pt] at (-4.67,-4) (B1) {};
\node[inner sep=0pt] at (-0.73,-4) (B2) {};
\node[inner sep=0pt] at (1.24,-3.5) (C1) {};
\node[inner sep=0pt] at (2.57,-3.5) (C2) {};
\node[inner sep=0pt] at (1.24,-4) (D1) {};
\node[inner sep=0pt] at (6.8,-4) (D2) {};
\node[inner sep=0pt] at (9,-3.5) (sc1) {67.8};
\draw[<->, >=stealth] (A1) -- node [midway,above,align=center] {Valeria} (A2);
\draw[<->, >=stealth] (B1) -- node [midway,above,align=center] {legge} (B2);
\draw[<->, >=stealth] (C1) -- node [midway,above,align=center] {il} (C2);
\draw[<->, >=stealth] (D1) -- node [midway,above,align=center] {giornale} (D2);
\node[align=left] at (-10,-5.5) (0w) {\textbf{Proper:}};
\node[inner sep=0pt] at (-6.95,-6.1) (F1) {};
\node[inner sep=0pt] at (-2.06,-6.1) (F2) {};
\node[inner sep=0pt] at (-4.86,-5.2) (G1) {};
\node[inner sep=0pt] at (0.47,-5.2) (G2) {};
\node[inner sep=0pt] at (2.57,-5.5) (H1) {};
\node[inner sep=0pt] at (2.95,-5.5) (H2) {};
\node[inner sep=0pt] at (0.09,-6) (I1) {};
\node[inner sep=0pt] at (6.8,-6) (I2) {};
\node[inner sep=0pt] at (9,-5.5) (sc2) {75.2};

\draw[<->, >=stealth] (F1) -- node [midway,above,align=center] {Valeria} (F2);
\draw[<->, >=stealth] (G1) -- node [midway,above,align=center] {legge} (G2);
\draw[<->, >=stealth] (H1) -- node [midway,above,align=center] {il} (H2);
\draw[<->, >=stealth] (I1) -- node [midway,above,align=center] {\ \ \ \ \ \ \ \ \  giornale} (I2);
\node at (-10,-7.5) (0w) {\textbf{Attention:}};
\node[inner sep=0pt] at (-8.6,-8) (Q1) {};
\node[inner sep=0pt] at (-6.57,-8) (Q2) {};
\node[inner sep=0pt] at (-6.57,-7.5) (W1) {};
\node[inner sep=0pt] at (-5.55,-7.5) (W2) {};
\node[inner sep=0pt] at (-5.55,-8) (E1) {};
\node[inner sep=0pt] at (-3.01,-8) (E2) {};
\node[inner sep=0pt] at (-3.01,-7.5) (R1) {};
\node[inner sep=0pt] at (1.04,-7.5) (R2) {};
\node[inner sep=0pt] at (1.04,-8) (T1) {};
\node[inner sep=0pt] at (6.12,-8) (T2) {};
\node[inner sep=0pt] at (6.12,-7.5) (Y1) {};
\node[inner sep=0pt] at (7.14,-7.5) (Y2) {};
\node[inner sep=0pt] at (9,-7.5) (sc3) {6.0};

\draw[<->, >=stealth] (Q1) -- node [midway,above,align=center] {il} (Q2);
\draw[<->, >=stealth] (W1) -- node [midway,above,align=center] {legge} (W2);
\draw[<->, >=stealth] (E1) -- node [midway,above,align=center] {il} (E2);
\draw[<->, >=stealth] (R1) -- node [midway,above,align=center] {giornale} (R2);
\draw[<->, >=stealth] (T1) -- node [midway,above,align=center] {Valeria} (T2);
\draw[<->, >=stealth] (Y1) -- node [midway,above,align=center] {giornale} (Y2);


\end{tikzpicture}